\documentclass{article}
\usepackage[final]{neurips_2023}  

\usepackage[utf8]{inputenc}
\usepackage[T1]{fontenc}
\usepackage{amsmath}
\usepackage{amsfonts}
\usepackage{amssymb}
\usepackage{graphicx}
\usepackage{hyperref}
\usepackage{booktabs}
\usepackage{algorithm}
\usepackage{algpseudocode}  
\usepackage{lipsum}
\usepackage{tikz}
\usetikzlibrary{shapes.geometric, arrows.meta, positioning}
\usepackage{caption}
\usepackage{float}
\usepackage{enumitem}
\usepackage{graphicx}
\usepackage{subcaption}
\usepackage{parskip}

\makeatletter
\renewcommand{\@noticestring}{}
\makeatother

\title{Unraveling Text Generation in LLMs: A Stochastic Differential Equation
Approach}

\author{
YUKUN ZHANG \\
The Chinese University Of Hong Kong\\
\texttt{215010026@link.cuhk.edu.cn}
}

\begin{document}

\maketitle

\begin{abstract}
This paper explores the application of Stochastic Differential Equations (SDE) to interpret the text generation process of Large Language Models (LLMs) such as GPT-4. Text generation in LLMs is modeled as a stochastic process where each step depends on previously generated content and model parameters, sampling the next word from a vocabulary distribution. We represent this generation process using SDE to capture both deterministic trends and stochastic perturbations. The drift term describes the deterministic trends in the generation process, while the diffusion term captures the stochastic variations. We fit these functions using neural networks and validate the model on real-world text corpora. Through numerical simulations and comprehensive analyses, including drift and diffusion analysis, stochastic process property evaluation, and phase space exploration, we provide deep insights into the dynamics of text generation. This approach not only enhances the understanding of the inner workings of LLMs but also offers a novel mathematical perspective on language generation, which is crucial for diagnosing, optimizing, and controlling the quality of generated text.
\end{abstract}

\section{Introduction}

\subsection{Background}
Large Language Models (LLMs), such as GPT-4, have revolutionized the field of Natural Language Processing (NLP) by demonstrating the ability to generate coherent and contextually relevant text. These models are typically based on deep learning architectures, such as transformers, which leverage large amounts of training data to learn intricate patterns in natural language. The success of LLMs in various applications, including text completion, translation, summarization, and question answering, has underscored their significance and utility.Despite their remarkable performance, LLMs operate as black-box models, making it challenging to interpret their decision-making processes. Understanding how these models generate text is crucial for several reasons. First, interpretability can enhance trust and transparency, allowing users to understand the rationale behind the generated content. Second, it can help identify and mitigate biases embedded in the model, which is essential for ensuring fairness and ethical use. Third, interpretability facilitates debugging and optimizing models, leading to improved performance and robustness.\\

Text generation in LLMs is inherently a sequential decision process where each step depends on previously generated content and model parameters. This process can be viewed as a stochastic process due to the inherent randomness and dynamic nature of generating text. To better understand and interpret this generation process, we propose modeling it using Stochastic Differential Equations (SDE).SDEs provide a mathematical framework to describe systems influenced by both deterministic trends and stochastic perturbations. By applying SDEs to model the text generation process, we aim to capture both the deterministic aspects of the model's learned language patterns (drift) and the random variations introduced during generation (diffusion). This approach allows for a nuanced interpretation of the generation process, offering insights into how LLMs produce text and highlighting areas where randomness plays a significant role.\\

To address these challenges, there is a pressing need to understand the generation process of LLMs. This understanding involves deciphering how these models utilize previously generated content and model parameters to sample the next word in a sequence, which inherently involves randomness and dynamic decision-making.

\subsection{Related Work}
\subsubsection{Stochastic Processes in NLP}
Stochastic processes have been widely utilized in natural language processing (NLP) to model the inherent randomness and variability of human language. These processes provide a comprehensive mathematical framework for capturing the probabilistic nature of text generation and interpreting various linguistic phenomena \cite{martinez2020light,harrison2017toward,drovo2019named,malik2017urdu}. One of the earliest and most notable applications of stochastic processes in NLP is the use of Hidden Markov Models (HMMs) for tasks such as speech recognition and part-of-speech (POS) tagging, where sequences of words are modeled as Markov processes \cite{zhang2020generating,martinez2020light,gehrmann2019improving,luo2016text,yang2018automatically}.

In addition, stochastic differential equations (SDEs) have been investigated to model the continuous evolution of language dynamics\cite{wang2023languagemodelingstochasticprocesses,debowski2020information}.These models offer insights into how language evolves over time and how various factors influence the generation process. By combining stochastic processes with deep learning models, researchers aim to capture both deterministic and stochastic elements of language, thus enhancing the robustness and variability in text generation.

\subsubsection{Interpretability in LLMs}

Interpretability in Large Language Models (LLMs) has been a major research focus, with early efforts centered on analyzing activations and attention maps to understand which parts of the input sequence the model focuses on during processing. These methods offered initial insights into transformer models by visualizing token interactions, but primarily captured static snapshots of model behavior. To overcome this limitation, more advanced techniques have been developed, such as model distillation, which simplifies complex models for better interpretability, and feature attribution methods like Integrated Gradients and SHAP, which assign importance scores to input features driving model outputs \cite{yang2023local, modarressi2023decompx, enguehard2023sequential}.

Recent advancements include linear decomposition for ReLU-activated transformers \cite{yang2023local}, DecompX for multi-layer transformers \cite{modarressi2023decompx}, and Sequential Integrated Gradients (SIG), which maintains sentence meaning while calculating word importance \cite{enguehard2023sequential}. Techniques such as interpretable autoprompting (iPrompt) \cite{singh2023explaining}, the interpretable autonomous driving system DriveGPT4 \cite{xu2023drivegpt4}, and Language Guided Bottlenecks (LaBo) \cite{yang2023language} further enhance LLM transparency. Interactive visualization tools, like InterpreT \cite{lal2021interpret}, provide additional means for analyzing transformer models.

Mechanistic explanations aim to uncover deeper insights into the inner workings of LLMs. Tools developed by Stan et al. \cite{stan2024lvlm} allow for interactive exploration of vision-language models, while Wang et al. \cite{wang2022interpretability} investigate how GPT-2 uses attention heads for indirect object identification. Todd et al. \cite{todd2023function} introduced function vectors, which transport task representations across contexts, and Creswell et al. \cite{creswell2022selection} proposed the Selection-Inference framework for generating causal reasoning steps in LLMs. Luo et al. \cite{luo2023reasoning} integrated LLMs with knowledge graphs for more faithful reasoning, while Chen et al. \cite{chen2024selfie} introduced SelfIE for interpreting embeddings.

Despite these advances, many existing methods still focus on static analysis and struggle to capture the dynamic and stochastic nature of LLM-generated text. To address this gap, we propose the use of Stochastic Differential Equations (SDEs) to model the text generation process in LLMs, capturing both deterministic trends (drift term) and stochastic variations (diffusion term). This approach provides a more dynamic and nuanced understanding of how LLMs generate text, improving interpretability, model diagnosis, and optimization \cite{huang2023can}.

\subsubsection{Motivation and Objectives}
Existing methods for explaining the text generation process of Large Language Models (LLMs) have notable limitations, particularly in capturing the complex balance between deterministic patterns and stochastic variations inherent in language generation. These methods often fail to provide a comprehensive understanding of the underlying mechanisms, focusing too narrowly on static analysis or deterministic elements while neglecting the significant role of randomness. To address these shortcomings, we propose the use of Stochastic Differential Equations (SDEs) as a novel explanatory framework. The primary objective of this study is to develop a mathematical model of the LLMs' text generation process using SDEs, thereby enhancing the interpretability and transparency of these models.

The innovation of our approach lies in the application of SDEs to explain the text generation process in LLMs. SDEs offer a unique advantage by simultaneously modeling the deterministic trends and random perturbations within the text generation process. This dual capability allows for a more nuanced understanding of how LLMs produce coherent and contextually relevant text while accounting for the inherent variability and unpredictability of language. By leveraging SDEs, our method addresses the deficiencies of existing interpretative approaches, providing a more robust and mathematically rigorous explanation of LLMs' internal mechanisms.

\subsection{Our contributions and the structure of paper}
The key contributions of this research include the introduction of a new mathematical framework based on SDEs for modeling the text generation process of LLMs, offering fresh insights into the internal workings of these models. Our findings hold significant theoretical and practical implications for the field of AI interpretability. The SDE-based framework not only advances the theoretical understanding of LLMs but also has the potential to support the development of safer and more controllable AI systems. By improving the interpretability of LLMs, this approach contributes to the broader goal of aligning AI systems with human values and ensuring their safe deployment across various applications.
This paper investigates how Stochastic Differential Equations (SDEs) can be used to model and enhance text generation in Large Language Models (LLMs). The structure is as follows:
\begin{itemize}
    \item \textbf{Section 2: Theory} 
    We introduce the fundamentals of SDEs and their application to LLMs. This section explains how drift and diffusion terms within SDEs capture the deterministic and stochastic elements of text generation, guiding the evolution of word embeddings and ensuring the production of coherent and contextually relevant text.

    \item \textbf{Section 3: Theoretical Analysis} 
    This section provides a rigorous analysis of the SDE-based model for text generation. We prove the existence and uniqueness of solutions, ensuring the model’s reliability. Additionally, we conduct a stability analysis to verify that the generated text remains coherent over time, and we explore the statistical properties of the model’s output through moment analysis.

    \item \textbf{Section 4: Experiment} 
    We empirically test the SDE framework in the context of LLM text generation. The experiment evaluates the model’s ability to balance coherence and variability, demonstrating how well the SDE approach captures the complexities of language generation in practice.

    \item \textbf{Appendix A: Proofs and Mathematical Details} 
    Detailed mathematical proofs supporting the theoretical analysis are provided here, including proofs of existence, uniqueness, and stability, reinforcing the robustness of the SDE-based approach.
\end{itemize}

\section{Theory}
\subsection{Fundamentals of Stochastic Differential Equations}

\subsubsection{Basic Definitions}
Stochastic Differential Equations (SDEs) are mathematical models that describe systems influenced by both deterministic trends and random perturbations. These equations are widely used in fields such as physics, finance, and biology to model the dynamics of systems over time under uncertainty. Recently, SDEs have also been applied in natural language processing (NLP) to model the complexity and variability inherent in language generation.

An SDE is typically formulated as:

\[
dX(t) = \mu(X(t), t)dt + \sigma(X(t), t)dW(t)
\]

where:
\begin{itemize}
    \item \( X(t) \) represents the state variable at time \( t \).
    \item \( \mu(X(t), t) \) is the drift term, capturing the predictable trends in the system.
    \item \( \sigma(X(t), t) \) is the diffusion term, modeling the random perturbations.
    \item \( dW(t) \) represents the Wiener process increment, the source of randomness in the system.
\end{itemize}

In simpler terms, the drift term \( \mu(X(t), t) \) guides the system along a logical path, while the diffusion term \( \sigma(X(t), t) \) introduces necessary randomness, capturing the unpredictable fluctuations.

By integrating SDEs into NLP, particularly in text generation, we can model the interplay between deterministic language structures and stochastic variations due to context and user interactions. For Large Language Models (LLMs), \( X(t) \) could represent the embedding of the generated text at time \( t \), \( \mu(X(t), t) \) could guide deterministic trends, and \( \sigma(X(t), t) \) could introduce the necessary randomness in word selection.

\subsubsection{State Variable in Text Generation}

In the context of Large Language Models (LLMs), the state variable \( X(t) \) is crucial for representing the generated text at each time step \( t \). Specifically, \( X(t) \) is defined as the word embedding of the token generated at time \( t \).

\paragraph{Definition of \( X(t) \)}
Word embeddings are dense vector representations that capture the meaning and syntactic roles of words. For instance, when generating the sentence “Hello world”, each word like “Hello” or “world” would have its own vector \( X(t) \), representing its meaning in a high-dimensional space. Formally, this can be represented as:

\[
X(t) \in \mathbb{R}^d
\]

where \( d \) is the dimensionality of the embedding space, and each component of \( X(t) \) represents a different aspect of the word's meaning or context.

\paragraph{Dimensionality and Structure}
The dimensionality \( d \) of \( X(t) \) depends on the embedding model, such as Word2Vec, GloVe, or BERT, and typically ranges from 100 to 1024 dimensions. These embeddings are trained to capture relationships between words, where words with similar meanings are placed closer together in this high-dimensional space. For example, in BERT, the dimensionality allows for a more nuanced understanding of word context, improving text generation's coherence and relevance.

\paragraph{Role in SDE Framework}
In the SDE framework for text generation, \( X(t) \) evolves over time, influenced by the drift term \( \mu(X(t), t) \), which drives the deterministic trends, and the diffusion term \( \sigma(X(t), t) \), which introduces stochastic variations. For example, as the model generates a sentence, \( X(t) \) shifts based on learned language patterns and introduces variability to avoid repetitive structures.This evolution allows the model to generate a sequence of embeddings that can be decoded into coherent and contextually appropriate text. The use of word embeddings ensures that the generated text remains semantically coherent and syntactically correct, as these embeddings encapsulate rich information about word meanings and contexts.In summary, \( X(t) \) represents the word embedding of the generated token at time \( t \). Its dimensionality and structure are vital for capturing the text's semantic and syntactic properties, playing a central role in the SDE-based text generation process in LLMs.

\subsubsection{Drift and Diffusion Terms in LLMs}

In the context of Large Language Models (LLMs), the Stochastic Differential Equation (SDE) framework employs both drift and diffusion terms to guide the text generation process. These terms work together to ensure that the generated text is both coherent and creatively diverse, capturing the deterministic and stochastic aspects of language.

\paragraph{Drift Term \( \mu(X(t), t) \)}
The drift term \( \mu(X(t), t) \) is responsible for the deterministic trends in the evolution of the state variable \( X(t) \), which represents the word embedding at time \( t \). Formally, it is expressed as:

\[
\mu(X(t), t) = f_{\mu}(X(t), t)
\]

The drift term, represented by \( f_{\mu} \), is a neural network or a parameterized function that maps the current state \( X(t) \) and time \( t \) to a vector in the same space as \( X(t) \). This term guides the model to generate text that follows logical and coherent patterns, incorporating elements such as contextual coherence, ensuring the generated words fit logically within the preceding text; language structure, embedding grammatical rules to maintain proper syntax and word order; thematic consistency, maintaining consistency in themes or arguments over extended text sequences; and predictive accuracy, reflecting the model's learned patterns to guide each word generation step based on previous context.

In the SDE framework, the drift term is crucial for driving the state variable \( X(t) \) along a predictable path, ensuring the text remains contextually relevant and grammatically correct.

\paragraph{Diffusion Term \( \sigma(X(t), t) \)}
Complementing the drift term, the diffusion term \( \sigma(X(t), t) \) introduces stochastic variability into the text generation process. It is defined as:

\[
\sigma(X(t), t) = f_{\sigma}(X(t), t)
\]

The diffusion term, represented by \( f_{\sigma} \), is another neural network or parameterized function that modulates the magnitude of random fluctuations, enabling the model to capture the inherent randomness and variability of human language. This term plays several key roles: it models variability by allowing random variations around deterministic trends, explores novel phrases by introducing randomness, handles uncertainty by generating a distribution of possible next states in ambiguous contexts, and prevents overfitting by ensuring the model does not become overly deterministic, thereby enhancing its generalization to new data.

The diffusion term is essential for capturing the subtle nuances and creative aspects of language, allowing the model to generate text that is not only coherent but also varied and engaging.

\paragraph{Integrated Role of Drift and Diffusion}
Together, the drift and diffusion terms in the SDE framework ensure that LLMs generate text that is both semantically coherent and creatively diverse. The drift term guides the text along learned patterns, while the diffusion term introduces the necessary randomness to explore new and varied linguistic possibilities. This balance is key to producing text that reflects both the structured nature of language and the unpredictability inherent in human communication.

In summary, the drift term \( \mu(X(t), t) \) and the diffusion term \( \sigma(X(t), t) \) are integral to the SDE-based text generation process in LLMs. While the drift term captures the deterministic trends essential for coherent text, the diffusion term introduces variability, enhancing the creativity, robustness, and generalization ability of the generated outputs.

\subsection{SDE Formulation for LLM Text Generation}

\subsubsection{Formulation of the SDE}

The Stochastic Differential Equation (SDE) formulation for text generation in Large Language Models (LLMs) integrates both deterministic and stochastic components to effectively model the dynamics of language. This SDE models the evolution of word embeddings over time, enabling the generation of coherent and contextually appropriate text.

\paragraph{Specific SDE for Text Generation}
The SDE used in this context is expressed as:

\[
dX(t) = \mu(X(t), t)dt + \sigma(X(t), t)dW(t)
\]

where:
\begin{itemize}
    \item \( X(t) \) represents the state variable at time \( t \), specifically the word embedding.
    \item \( \mu(X(t), t) \) is the drift term, capturing deterministic trends in the text generation process.
    \item \( \sigma(X(t), t) \) is the diffusion term, capturing stochastic variations.
    \item \( dW(t) \) denotes the increment of a Wiener process, modeling random noise.
\end{itemize}

\paragraph{Role of Each Term in Text Generation}
The drift term \( \mu(X(t), t) \) ensures that the generated text follows a coherent trajectory based on learned patterns, guiding the sequence towards syntactically and semantically correct constructs. The diffusion term \( \sigma(X(t), t) \) introduces variability, allowing the model to explore different word choices and structures, preventing repetitive and overly deterministic text. The Wiener process \( dW(t) \) adds random perturbations, simulating the natural variability in language, such as creative word choices or idiomatic expressions.

\paragraph{Parameterization Using Neural Networks}
Both the drift \( \mu(X(t), t) \) and diffusion \( \sigma(X(t), t) \) terms are parameterized using neural networks. These networks are trained to predict the next word embedding based on the current state, with the drift term guiding the deterministic evolution and the diffusion term introducing stochastic variations. The architecture typically involves input layers (current word embedding and time encoding), hidden layers (with non-linear activation functions), and output layers that produce the respective drift or diffusion vectors.

\paragraph{Integrated SDE Model}
Combining these elements, the integrated SDE model for text generation in LLMs is given by:

\[
dX(t) = \text{NN}_{\mu}(X(t), t; \theta_{\mu})dt + \text{NN}_{\sigma}(X(t), t; \theta_{\sigma})dW(t)
\]

This formulation leverages the power of neural networks to capture both deterministic and stochastic aspects of language generation, enabling the generation of diverse and creative text while maintaining coherence.

\paragraph{Training the Neural Networks}
The neural networks for \( \mu(X(t), t) \) and \( \sigma(X(t), t) \) are trained using a combination of supervised and unsupervised learning techniques, optimizing the parameters to minimize prediction errors and accurately capture variability in the training data.

\subsubsection{Learning the SDE Parameters}

Estimating the parameters of the Stochastic Differential Equation (SDE) for text generation involves optimizing the weights of the neural networks that define the drift and diffusion terms. This section outlines the methods for parameter estimation and the optimization techniques employed to ensure accurate and efficient learning.

\paragraph{Estimation Methods}
The parameters \( \theta_{\mu} \) and \( \theta_{\sigma} \) for the neural networks \( \text{NN}_{\mu} \) and \( \text{NN}_{\sigma} \), corresponding to the drift term \( \mu(X(t), t) \) and the diffusion term \( \sigma(X(t), t) \) respectively, are learned through supervised training with labeled data.

\paragraph{Loss Functions}
The neural networks are trained using loss functions that capture both deterministic and stochastic aspects of text generation:

\begin{itemize}
    \item \textbf{Drift Term Loss:} The loss for the drift term \( \mu(X(t), t) \) is defined as the mean squared error (MSE) between the predicted next state and the actual next state:
    \[
    \mathcal{L}_{\mu} = \frac{1}{N} \sum_{i=1}^{N} \left\| X(t_i + \Delta t) - \left( X(t_i) + \mu(X(t_i), t_i) \Delta t \right) \right\|^2,
    \]
    where \( N \) is the number of training samples.

    \item \textbf{Diffusion Term Loss:} The loss for the diffusion term \( \sigma(X(t), t) \) is designed to align the statistical properties of the predicted distribution with the actual distribution, commonly using the Kullback-Leibler (KL) divergence:
    \[
    \mathcal{L}_{\sigma} = \frac{1}{N} \sum_{i=1}^{N} \text{KL}\left( P_{data}(X(t_i + \Delta t) | X(t_i)) \parallel P_{model}(X(t_i + \Delta t) | X(t_i)) \right),
    \]
    where \( P_{data} \) and \( P_{model} \) represent the data distribution and model distribution, respectively.
\end{itemize}

In summary, learning the SDE parameters involves training neural networks to capture the complex, non-linear relationships in the text generation process. This is achieved through a structured training process that includes data preparation, precise loss function definition, and gradient-based optimization techniques.

\section{Theoretical Analysis of the Model}

The theoretical analysis of Stochastic Differential Equations (SDEs) in the context of Large Language Models (LLMs) is crucial for ensuring that these models generate stable, reliable, and contextually appropriate text over extended sequences. This section provides a comprehensive overview of the existence and uniqueness of solutions, stability analysis, and moment analysis, which are foundational for the design and implementation of robust LLMs.

\subsection{Existence and Uniqueness of Solutions}

The existence and uniqueness of solutions to the SDE governing text generation in LLMs are critical for ensuring stable and reliable outputs. The key conditions for ensuring these properties are Lipschitz continuity and linear growth for the drift term \( \mu(X(t), t) \) and the diffusion term \( \sigma(X(t), t) \). Specifically, these conditions can be formulated as:

\subsubsection{Lipschitz Continuity}
Both \( \mu(X(t), t) \) and \( \sigma(X(t), t) \) must satisfy a Lipschitz condition, meaning there exists a constant \( K \) such that for all \( X_1, X_2 \) and \( t \):
\[
\| \mu(X_1, t) - \mu(X_2, t) \| + \| \sigma(X_1, t) - \sigma(X_2, t) \| \leq K \| X_1 - X_2 \|
\]

\subsubsection{Linear Growth Condition}
Both \( \mu(X(t), t) \) and \( \sigma(X(t), t) \) must also satisfy a linear growth condition, where a constant \( C \) exists such that for all \( X \) and \( t \):
\[
\| \mu(X(t), t) \|^2 + \| \sigma(X(t), t) \|^2 \leq C (1 + \| X(t) \|^2)
\]

Under these conditions, the Picard-Lindelöf theorem guarantees the existence and uniqueness of the solution to the SDE, ensuring that the state evolution of the LLM is well-defined and predictable.

\subsection{Stability Analysis}

Stability analysis is essential for understanding the long-term behavior of the SDE solutions and ensuring that the generated text remains coherent over extended sequences. The stability of these solutions can be analyzed using Lyapunov functions, which help determine whether the system's state remains bounded and converges to a desired equilibrium.

\subsubsection{Lyapunov Function Method}
A Lyapunov function \( V(X(t)) \) is a scalar function used to assess the stability of an equilibrium point. For the SDE, the Lyapunov function is typically positive definite, and its derivative along the trajectories of the SDE should be non-positive:
\[
\mathcal{L}V(X(t)) = \frac{\partial V}{\partial X} \mu(X(t), t) + \frac{1}{2} \text{Tr} \left( \sigma(X(t), t)^\top \frac{\partial^2 V}{\partial X^2} \sigma(X(t), t) \right) \leq 0
\]

If \( \mathcal{L}V(X(t)) < 0 \), the equilibrium point is asymptotically stable, meaning the solution will converge to the equilibrium as time progresses.

\subsection{Moment Analysis}

Moment analysis is critical for understanding the statistical properties of the solutions to the SDE. Using Itô's lemma, we can analyze the mean and variance of the state variable \( X(t) \), as well as higher-order moments.

\subsubsection{Mean and Variance Analysis Using Itô's Lemma}
The SDE is given by:
\[
dX(t) = \mu(X(t), t)dt + \sigma(X(t), t)dW(t).
\]

To analyze the mean \( m(t) = \mathbb{E}[X(t)] \) and variance \( v(t) = \mathbb{E}[(X(t) - m(t))^2] \), we use the following relationships:
\[
\frac{d}{dt} m(t) = \mathbb{E}[\mu(X(t), t)]
\]
\[
\frac{d}{dt} \mathbb{E}[X(t)^2] = 2 \mathbb{E}[X(t) \mu(X(t), t)] + \mathbb{E}[\sigma^2(X(t), t)].
\]

Assuming linear relationships, these equations can be solved to provide insights into how the mean and variance evolve over time.

\subsubsection{Higher-order Moments}
The dynamics of higher-order moments \( \mathbb{E}[X(t)^n] \) can also be derived using Itô's lemma:
\[
\frac{d}{dt} \mathbb{E}[X(t)^n] = n \mathbb{E}[X(t)^{n-1} \mu(X(t), t)] + \frac{n(n-1)}{2} \mathbb{E}[X(t)^{n-2} \sigma^2(X(t), t)].
\]
This analysis helps in understanding the tail behavior of the distribution of \( X(t) \) and its impact on text generation.

\subsection{Summary and Discussion}

This theoretical analysis provides a robust framework for understanding the behavior of SDEs in LLMs. The existence and uniqueness conditions ensure well-defined solutions, while stability analysis guarantees that these solutions remain meaningful over time. Moment analysis reveals the distributional properties, ensuring that the generated text is both coherent and diverse. Collectively, these insights are crucial for the design of LLMs that produce high-quality and reliable text across various applications.

Future work will focus on empirical validation and further refinement of these models based on real-world data and applications.

\section{Experiment}

The primary objective of this experiment is to explore the application of Stochastic Differential Equations (SDEs) in modeling the text generation process of Large Language Models (LLMs) like GPT-4. By doing so, we aim to better understand and potentially control the behavior of these models, which is crucial for aligning AI outputs with human values and ensuring safe AI systems. Modeling the stochastic nature of text generation provides insights into how deterministic trends and random perturbations influence the generated content, thereby addressing a core challenge in AI alignment: the unpredictability and potential misalignment of LLM outputs.

The experiment evaluates key metrics such as total loss, drift loss, and diffusion loss to assess the model's performance. The results indicate that the model effectively learns to balance coherence and variability, with the drift term capturing the general direction of text generation and the diffusion term managing randomness. Additionally, trajectory analysis reveals how well the model's predictions align with actual generated text, especially in handling complex language structures. The refined analysis of drift and diffusion provides insights into the strengths and limitations of the current LLM architecture, guiding future improvements for more accurate and contextually relevant text generation

\subsection{Data and Model Description}

\paragraph{Data}
The HelpSteer dataset, sourced from NVIDIA, was selected for its rich annotations and real-world applicability in evaluating AI-generated text. It provides a robust foundation for assessing text generation models, containing diverse prompts and responses annotated with quality metrics such as helpfulness, correctness, coherence, complexity, and verbosity. 

\paragraph{Model Description}
The proposed model leverages Stochastic Differential Equations (SDEs) to capture the nuanced dynamics of text generation in Large Language Models (LLMs). The architecture consists of two key components: a Drift Network, which predicts the general direction or trends in word sequences, and a Diffusion Network, which manages the random variations or uncertainties in the text generation process. Both networks are designed as multi-layer perceptrons (MLPs) aligned with the GPT-4 embedding size and are trained together to optimize a loss function. This loss function balances the model’s ability to accurately predict both the deterministic shifts and the inherent randomness in text generation. This innovative integration of SDEs with GPT-4's architecture provides a novel and mathematically grounded approach to improving the understanding and performance of LLMs in generating coherent and contextually relevant text.

\subsection{Experimental Results Analysis}

\subsubsection{Evaluation of Stochastic Differential Equation (SDE) Model}

\begin{figure}[htbp]
    \centering
    \includegraphics[width=\textwidth]{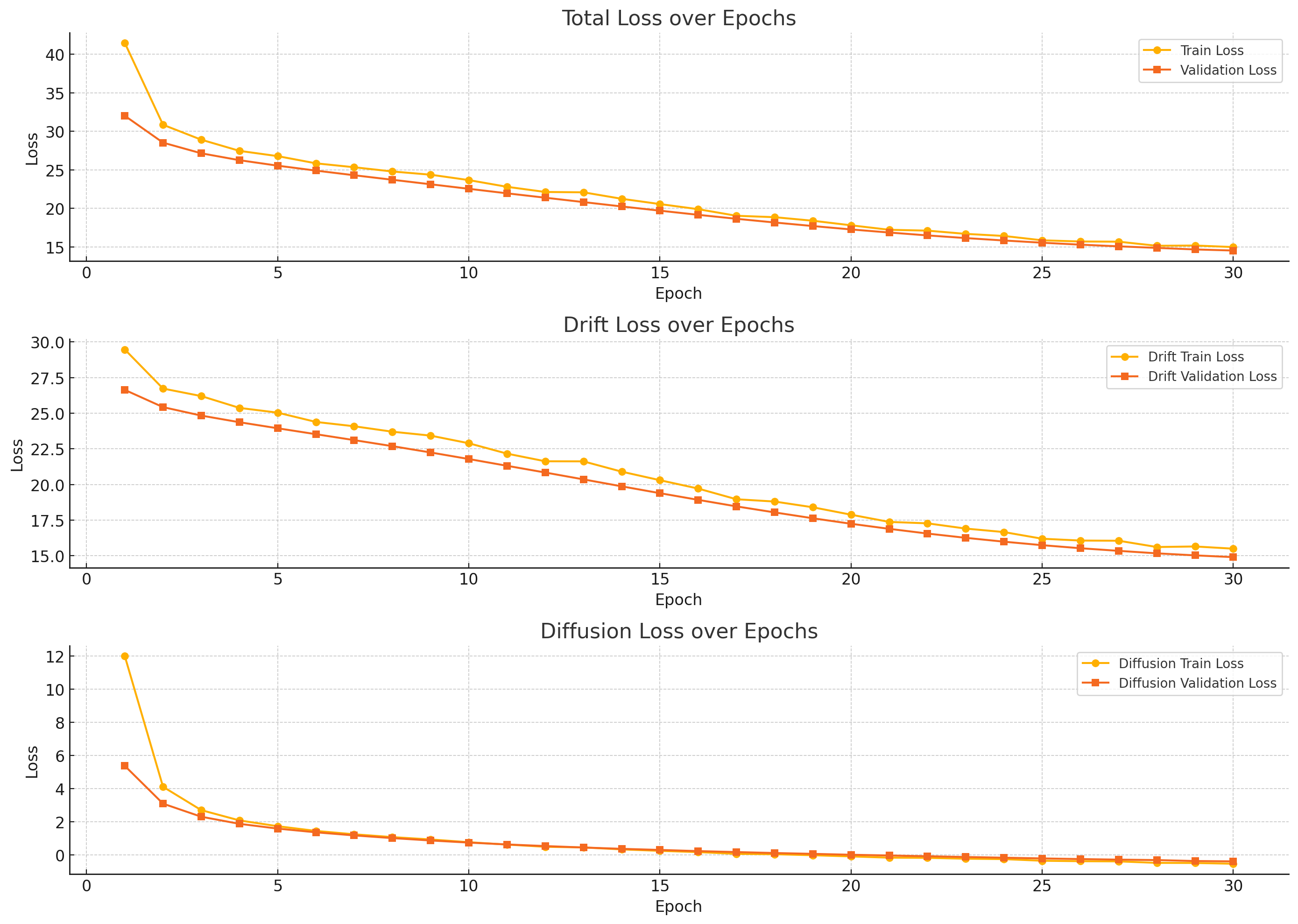}
    \caption{Training and Validation Losses over Epochs for the SDE Model. The top plot shows the total loss, the middle plot shows the drift loss, and the bottom plot shows the diffusion loss. The model shows a consistent decrease in losses over the epochs, indicating improved learning and stability.}
    \label{fig:sde_losses}
\end{figure}

In evaluating the Stochastic Differential Equation (SDE) model, several key metrics were utilized to assess its performance during the text generation process. These metrics include total loss, drift loss, and diffusion loss, each providing unique insights into the model’s learning dynamics:

\begin{itemize}
    \item \textbf{Total Loss}: Serves as a comprehensive indicator of the model's overall performance, combining both deterministic and stochastic elements of the text generation process.
    \item \textbf{Drift Loss}: Reflects the model’s ability to capture deterministic trends in language, indicating how well the model can predict the next word based on the context provided by previous tokens.
    \item \textbf{Diffusion Loss}: Measures the model's management of stochasticity or randomness, highlighting its effectiveness in balancing coherence with variability during text generation.
\end{itemize}

\textbf{Convergence and Dynamics of Loss Functions:} The evaluation revealed a consistent decrease in training and validation loss curves across all epochs, as illustrated in Figure \ref{fig:sde_losses}. The reduction in drift loss confirms the model's capability to capture the deterministic structure of language, while the stabilization or reduction in diffusion loss suggests a controlled reduction in randomness. This interplay between total loss, drift loss, and diffusion loss is crucial for understanding how the model balances coherence with variability.

\subsubsection{Trajectory Analysis}

\begin{figure}[htbp]
    \centering
    \includegraphics[width=0.8\textwidth]{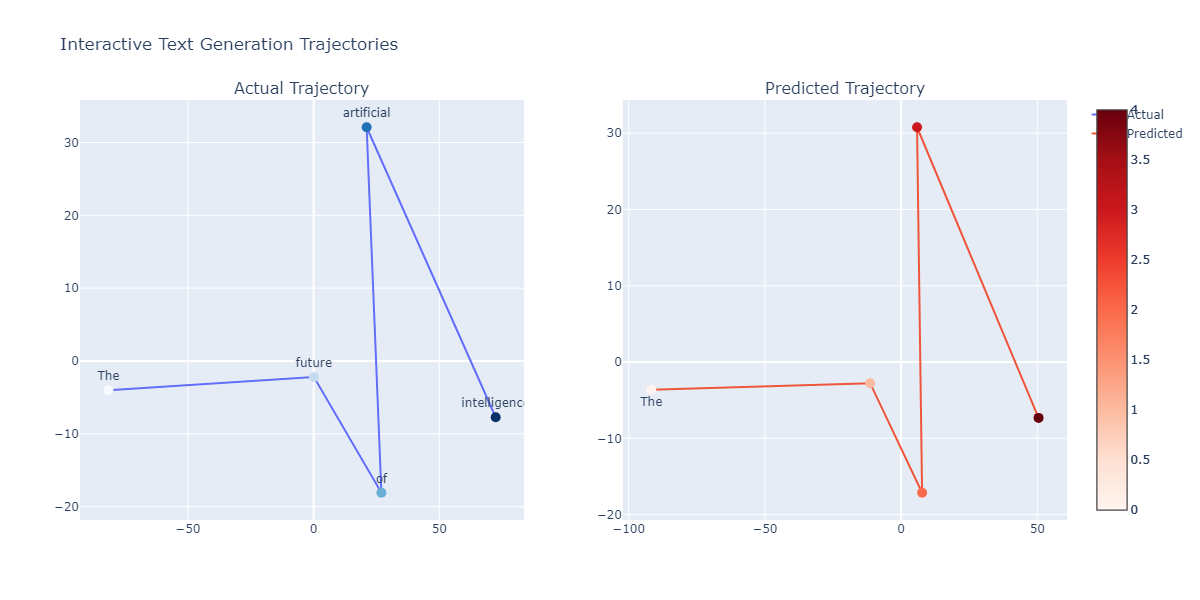}
    \caption{Actual vs Predicted Trajectories. The left plot shows the actual trajectory of the text generation, while the right plot presents the predicted trajectory. The comparison between these trajectories highlights the alignment and discrepancies in the model's predictions.}
    \label{fig:figure2}
\end{figure}

\begin{figure}[htbp]
    \centering
    \includegraphics[width=0.8\textwidth]{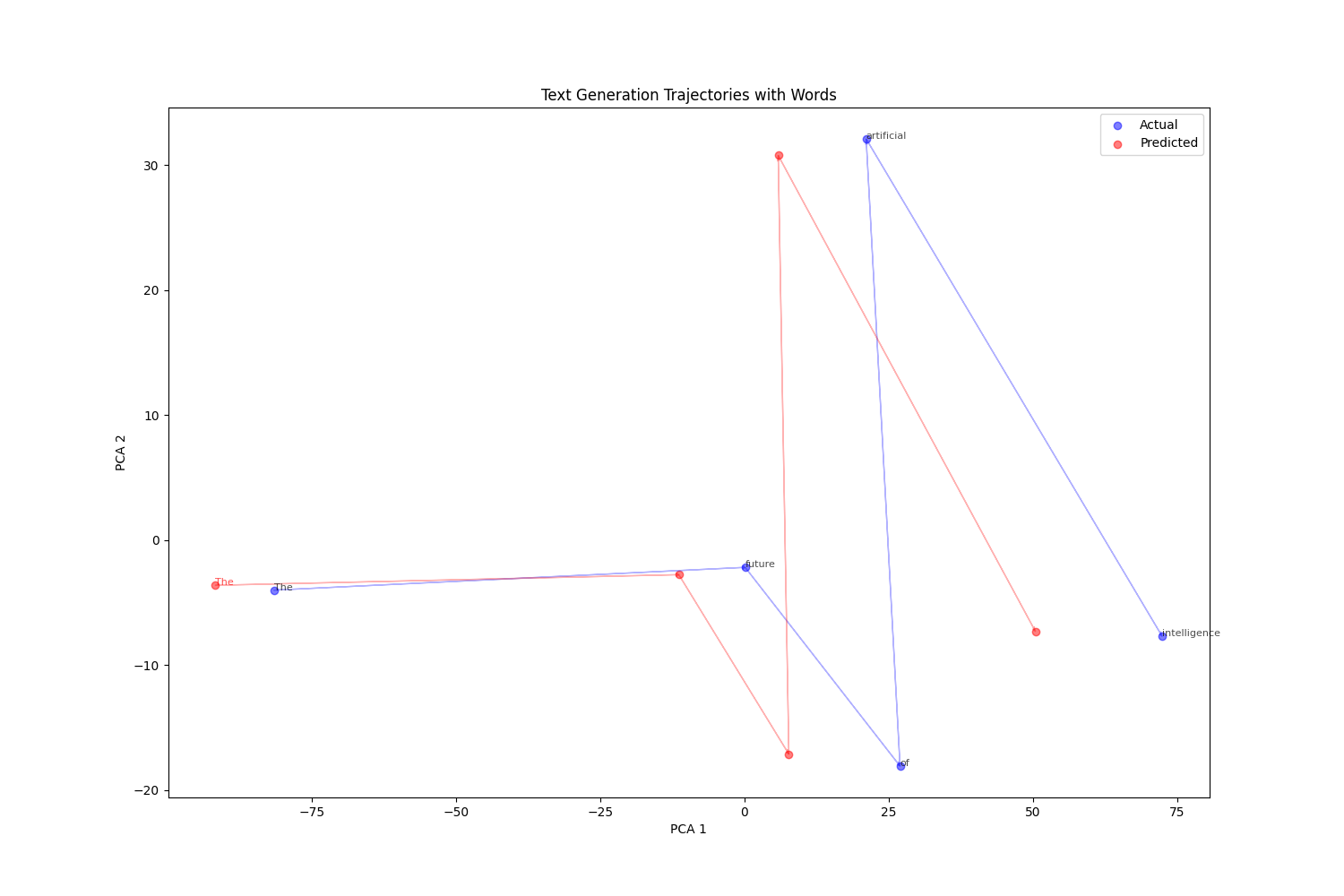}
    \caption{Text Generation Trajectories with Words. This plot visualizes the actual and predicted word trajectories during the text generation process. Each point represents a word in the sequence, showing the semantic movement in the PCA-reduced space. The alignment between the red (predicted) and blue (actual) paths illustrates the model's accuracy in capturing the semantic evolution.}
    \label{fig:figure3}
\end{figure}

The provided images (Figure \ref{fig:figure2} and Figure \ref{fig:figure3}) illustrate the actual and predicted text generation trajectories, as well as a comparison between the two. These visualizations are critical in understanding how well the model’s predicted paths align with the actual word sequences generated by the language model (LLM).

\textbf{Key Insights:}

\begin{itemize}
    \item \textbf{Trajectory Alignment:} The figures demonstrate the degree of alignment between the actual text generation trajectories and those predicted by the model. In both the actual vs. predicted trajectory plots and the PCA-based trajectory comparison, we observe that the model's predictions closely follow the actual trajectory during the initial stages of text generation. This indicates that the model's drift term effectively captures the general direction and flow of the generated text when dealing with simpler sentence structures.
    
    \item \textbf{Deviations in Complex Scenarios:} As the complexity of the text increases, such as in sentences involving multiple clauses or nuanced semantic transitions, the predicted trajectory starts to diverge more noticeably from the actual trajectory. This deviation suggests that while the model performs well in straightforward contexts, its ability to predict text generation paths diminishes as the linguistic complexity increases. These discrepancies highlight potential limitations in the drift term's capacity to fully encapsulate the intricacies of more complex language constructs.
    
    \item \textbf{Dimensionality Reduction via PCA:} The PCA-based visualization effectively reduces the dimensionality of the word embeddings, making it easier to observe the overall trajectory trends. This analysis confirms that the primary components of the trajectory are well captured by the model, but it also reveals the points at which the predicted trajectory diverges from the actual path, further emphasizing areas where model improvements are necessary.
\end{itemize}

\textbf{Conclusion:} The trajectory analysis underscores the strengths and limitations of the current LLM's text generation process. The model demonstrates a strong ability to predict text generation trajectories in simple linguistic contexts, where the predicted paths closely align with the actual generated sequences. However, as the complexity of the language increases, the model's predictions become less accurate, indicating a need for further refinement of the drift and diffusion terms within the stochastic differential equation (SDE) framework.

These findings suggest that enhancing the model's capacity to handle complex linguistic structures, possibly through more sophisticated drift term modeling or improved integration of contextual information, could lead to more accurate and coherent text generation. The trajectory analysis thus provides a critical diagnostic tool for guiding future developments in LLM architecture, ultimately contributing to the creation of more robust and versatile language models.

\subsubsection{Drift and Diffusion Analysis}

In this section, we analyze the trajectory of text generation by a large language model (LLM) through the lens of Stochastic Differential Equations (SDE), focusing on the key aspects of Drift and Diffusion. These concepts help us understand the underlying dynamics of the text generation process and the model's behavior over time.

\begin{figure}[htbp]
\centering
\includegraphics[width=0.45\textwidth]{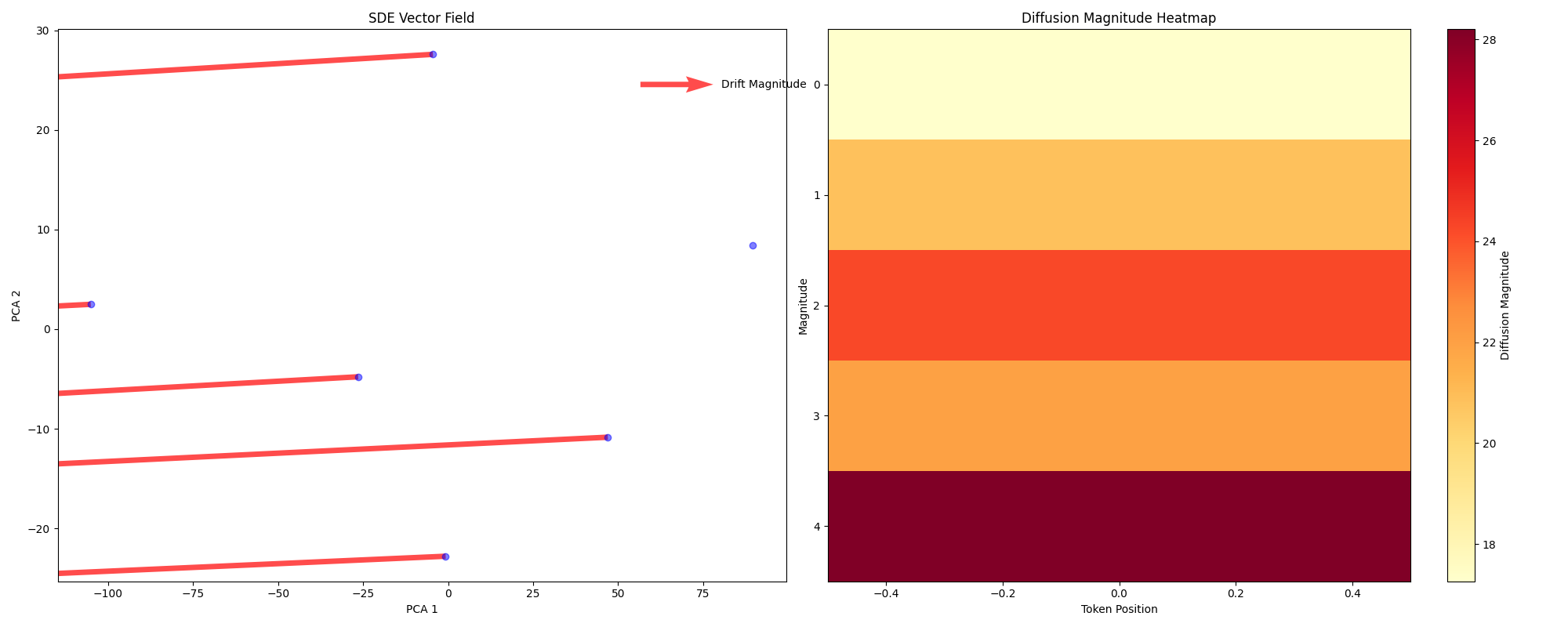}
\includegraphics[width=0.45\textwidth]{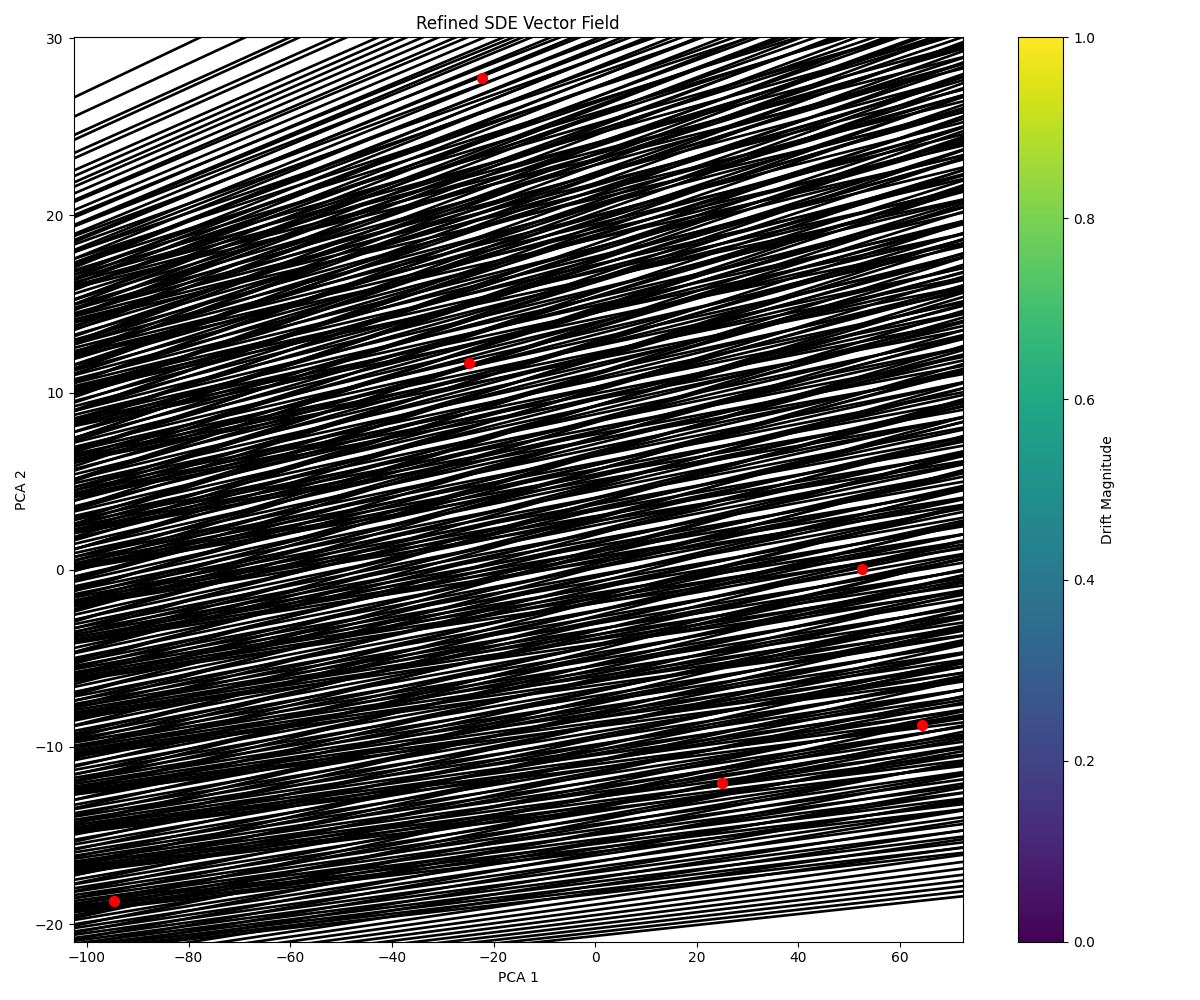}
\caption{(Left) SDE Vector Field illustrating the drift magnitude during the text generation process. (Right) Diffusion Magnitude Heatmap showing the diffusion effects across different token positions.}
\label{fig:figure4_8}
\end{figure}

\textbf{Drift Vector Field (Left Panel)}: The drift vector field represents the deterministic component of the text generation process. In the context of LLMs, the drift vectors indicate the predominant direction and magnitude of the model's predictions as it generates text. The arrows in the field illustrate how the model's internal state evolves as it transitions from one word to the next. A strong drift in a particular direction suggests that the model is confident in its trajectory, steadily moving towards a specific word or phrase in the latent semantic space. This deterministic movement can be thought of as the model's learned bias towards certain sequences of words, reflecting its training data and the patterns it has internalized.

\textbf{Diffusion Magnitude Heatmap (Right Panel)}: The diffusion magnitude represents the stochastic component, capturing the uncertainty or variability in the model's predictions. The heatmap shows how this uncertainty varies across different stages of the text generation process. High diffusion values correspond to regions where the model exhibits greater uncertainty, possibly due to ambiguous context or less frequent word sequences in the training data. These areas of high diffusion indicate points where the model's predictions are more likely to deviate from the expected path, potentially leading to more diverse or creative outputs. Conversely, low diffusion areas suggest more confident predictions with less variability, where the model's output is more predictable.

\begin{figure}[htbp]
\centering
\includegraphics[width=0.45\textwidth]{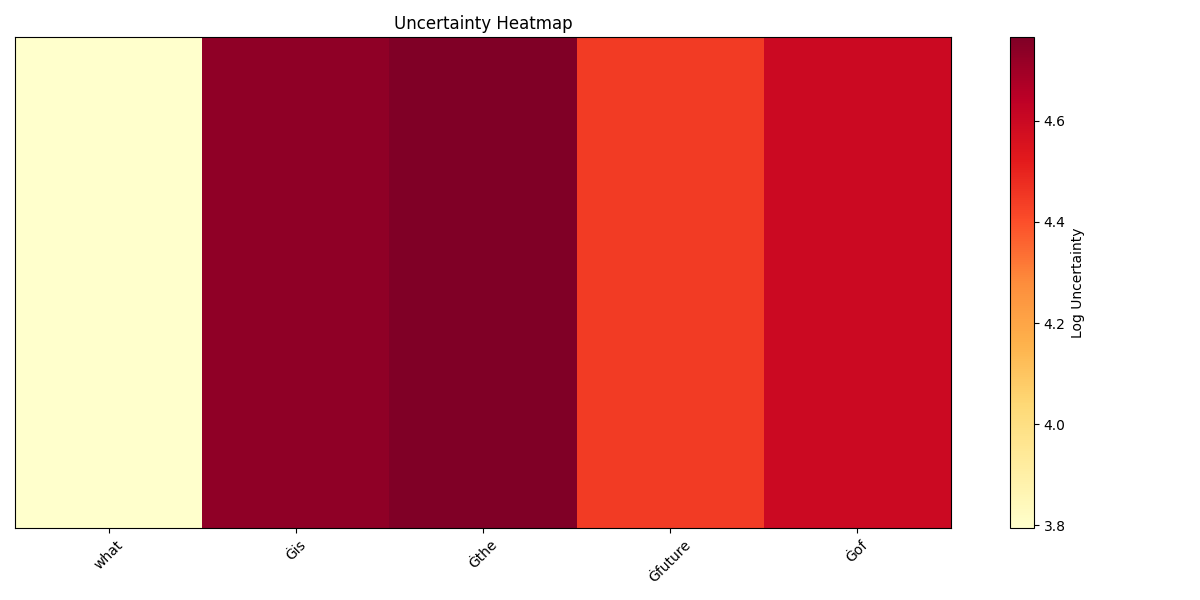}
\includegraphics[width=0.45\textwidth]{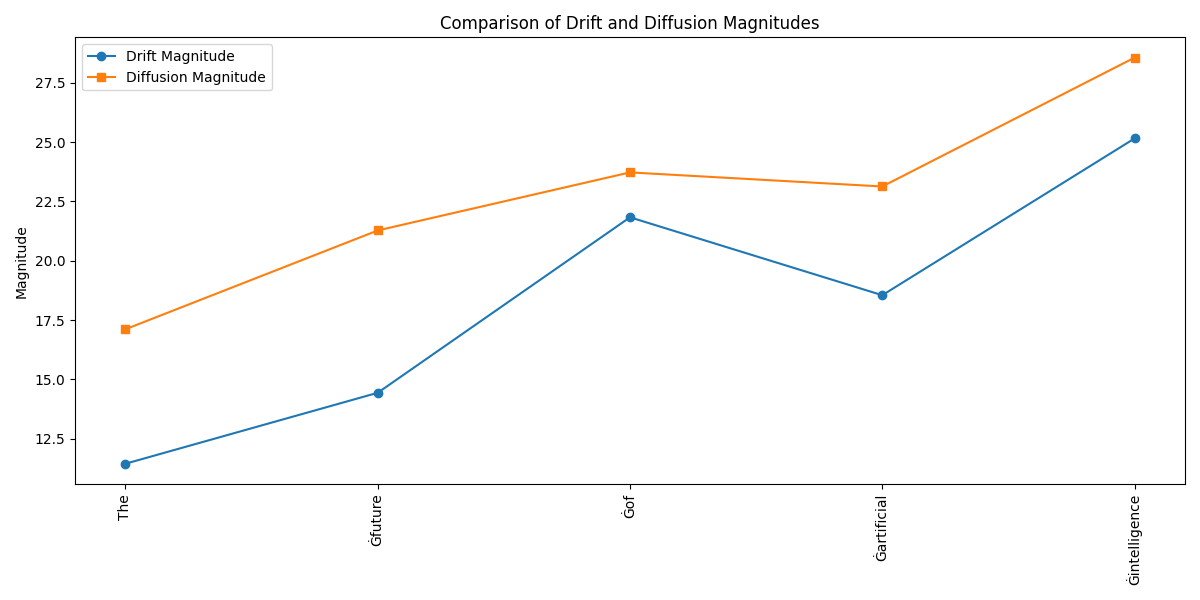}
\caption{(Left) Refined SDE Vector Field highlighting drift magnitude across various components. (Right) Uncertainty Heatmap demonstrating the uncertainty across different tokens during generation.}
\label{fig:figure9_10}
\end{figure}

\textbf{Refined Drift Vector Field (Left Panel)}: This figure presents a more detailed analysis of the drift vectors, allowing us to examine the model's behavior across different principal components in the latent space. By focusing on these principal components, we can observe the primary directions in which the model's state evolves as it generates text. The refined drift vectors provide insights into the model's decision-making process, highlighting the dominant semantic directions that guide text generation. This analysis can reveal the underlying structures and biases in the model's language generation process, helping to identify systematic trends or anomalies.

\textbf{Uncertainty Heatmap (Right Panel)}: The uncertainty heatmap complements the diffusion analysis by providing a log-scale visualization of the model's uncertainty across different generated words. This heatmap allows us to pinpoint specific words or phrases where the model exhibits high uncertainty, indicating potential areas of improvement or further training. By analyzing these uncertainty patterns, we can better understand the situations where the model struggles, such as generating coherent responses in complex contexts or dealing with ambiguous inputs. The log-scale also enhances the contrast, making it easier to identify critical points of uncertainty that could significantly impact the generated text's quality.

\textbf{Conclusion:} The drift and diffusion analyses provide a comprehensive view of the LLM's behavior during text generation. The drift vectors illustrate the deterministic paths that the model tends to follow, reflecting the patterns it has learned from the training data. On the other hand, the diffusion magnitudes and uncertainty heatmaps highlight the model's variability and the challenges it faces in predicting certain word sequences. Together, these analyses offer valuable insights into the strengths and weaknesses of the LLM, guiding future improvements in model architecture, training strategies, and text generation techniques.

This SDE-based analysis framework is crucial for understanding and enhancing the performance of large language models, particularly in scenarios where precise control over generated text is necessary. By dissecting the deterministic and stochastic elements of the generation process, we can develop more robust and reliable models that better meet the demands of real-world applications.

\subsubsection{Analysis of Word Importance and Attention Weights in Text Generation}

\begin{figure}[htbp]
    \centering
    \begin{subfigure}[b]{0.48\textwidth}
        \centering
        \includegraphics[width=\textwidth]{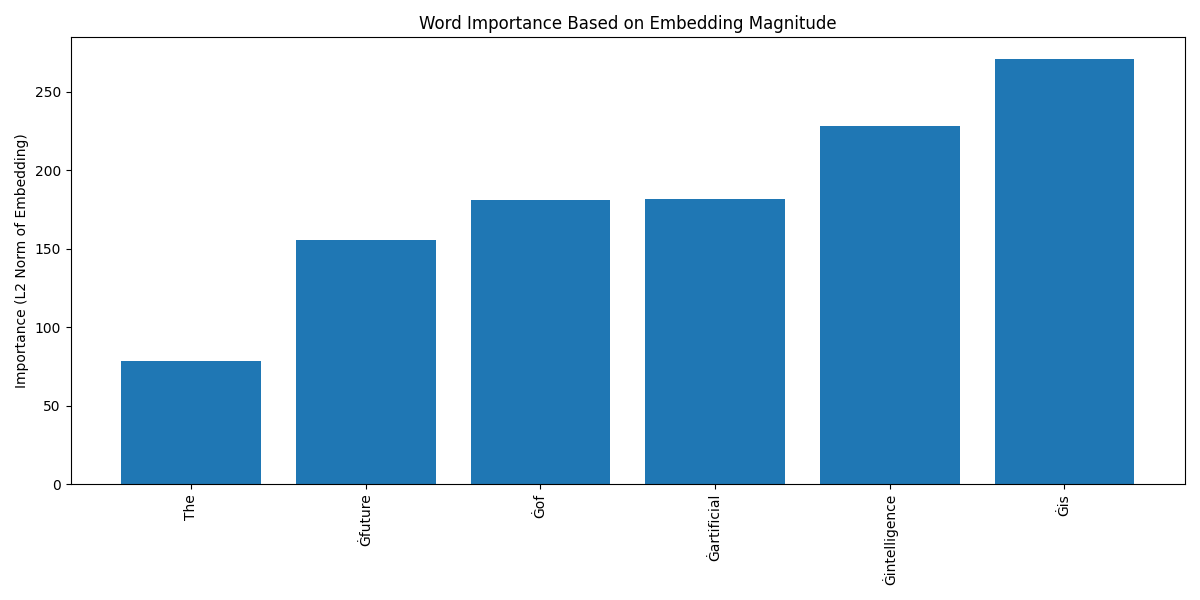}
        \caption{Word Importance Based on Embedding Magnitude}
        \label{fig:word_importance}
    \end{subfigure}
    \hfill
    \begin{subfigure}[b]{0.48\textwidth}
        \centering
        \includegraphics[width=\textwidth]{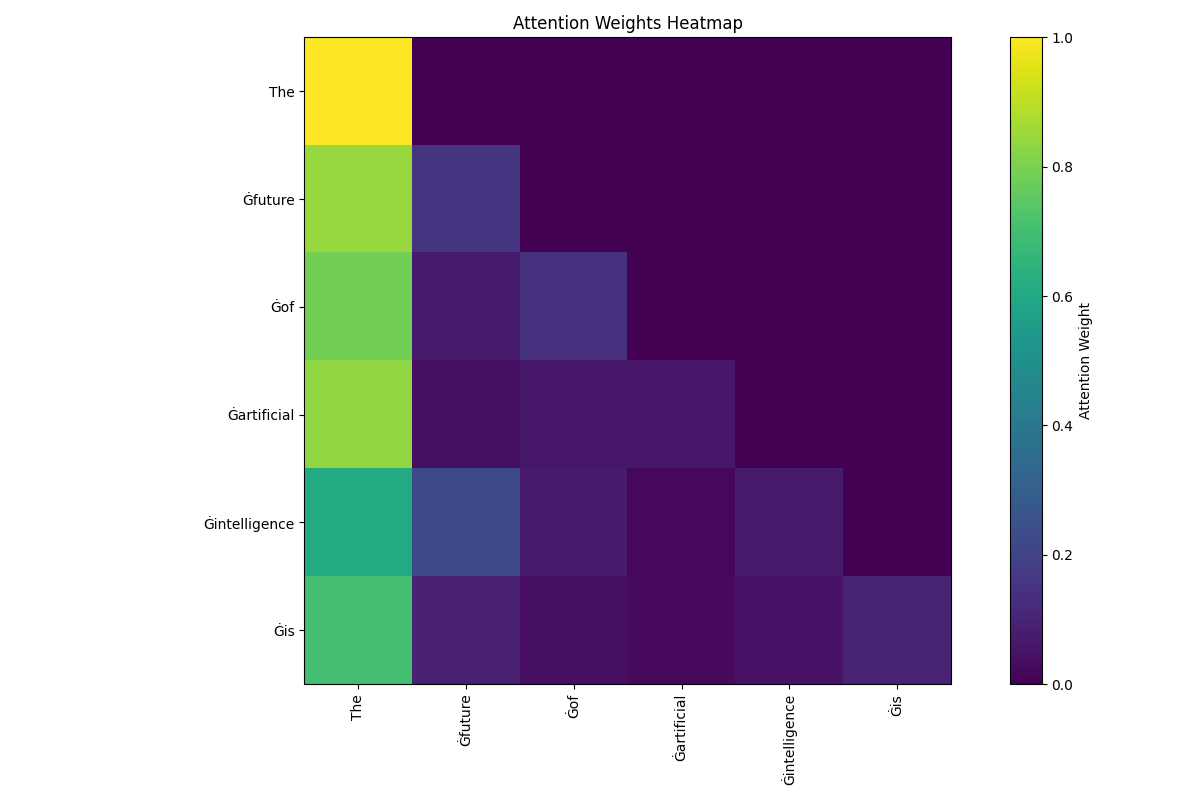}
        \caption{Attention Weights Heatmap}
        \label{fig:attention_weights}
    \end{subfigure}
    \caption{Analysis of Word Importance and Attention Weights in Text Generation}
    \label{fig:importance_attention_analysis}
\end{figure}

The analysis in Figures \ref{fig:word_importance} and \ref{fig:attention_weights} explores the significance of word importance and attention weights in understanding how a language model (LLM) such as GPT or BERT assigns relevance to different tokens during text generation.\\

Figure \ref{fig:word_importance} demonstrates the importance of various words based on the L2 norm of their embeddings. Certain words exhibit significantly higher magnitudes in the embedding space, indicating their influential role in shaping the context and meaning of the generated text. For example, tokens like "Gis" and "Gintelligence" show higher importance, suggesting they heavily influence subsequent tokens generated by the model. Such high-magnitude words are likely to dominate the semantic direction of the generated sequence, which is critical in tasks requiring precise control over content and style.

Figure \ref{fig:attention_weights} illustrates the distribution of attention weights across different tokens. This heatmap visually represents how the model allocates focus among words as it generates text. Words like "The" and "Gfuture" receive higher attention in specific layers, highlighting their contextual importance during certain stages of the sequence generation. Understanding these attention weights is crucial for diagnosing model behavior, as it reveals where the model places its "focus" and which tokens it prioritizes during word generation.

\subsection{Conclusion}
This experiment explored the application of Stochastic Differential Equations (SDEs) to model the text generation process of Large Language Models (LLMs), such as GPT-4. By integrating both deterministic (drift) and stochastic (diffusion) components, the SDE framework provided a deeper understanding of how LLMs generate coherent and contextually relevant text. Key metrics, including word importance and attention weights, were analyzed to evaluate the model's performance. The results showed that the SDE-based model effectively balances coherence and variability, capturing both the predictable trends and randomness inherent in language generation. The findings indicate the potential for further improvement in handling complex language structures and improving model robustness.

\section{Summary and Discussion}
This paper presented a novel approach using Stochastic Differential Equations (SDEs) to model text generation in Large Language Models (LLMs) like GPT-4. By incorporating drift terms to capture deterministic trends and diffusion terms to handle stochastic variations, our model offers a dynamic and flexible method for generating coherent and contextually relevant text. The results showed that the SDE-based approach effectively balances consistency with creativity, allowing the model to follow logical word sequences while introducing variability where necessary. Analysis of word importance and attention weights further revealed how key tokens influence text generation, aligning with the SDE framework’s deterministic and stochastic components.

While the model performed well in straightforward text generation, challenges emerged with more complex language structures, indicating the need for further refinement. Future research could focus on improving the drift and diffusion terms to better handle nuanced linguistic contexts and exploring applications of SDEs in other areas of natural language processing. These enhancements could lead to more robust and versatile language models capable of generating high-quality text across a wider range of tasks.

Looking ahead, there are several promising directions for future research. One avenue is to investigate the integration of our SDE-based interpretation method with other explainable AI techniques, such as attention visualization or feature attribution methods. This could provide a more comprehensive understanding of LLM behavior. Another important direction is to explore how the insights gained from our SDE model can be used to improve LLM training and generation processes. For instance, the drift and diffusion terms identified in our model could potentially be used to guide the development of more controlled and reliable text generation algorithms. Finally, extending this work to multi-modal language models, which incorporate both text and visual information, presents an exciting challenge that could lead to broader applications in AI interpretation and development.

\bibliography{reference} 
\bibliographystyle{plainnat} 

\appendix
\section{Appendix A: Additional Proofs}
In the proof section of this paper, we utilize the theory of Stochastic Differential Equations (SDEs) to explain the evolution of state variables in Large Language Models (LLMs) during the text generation process. By applying the Picard-Lindelöf theorem, we demonstrate the existence and uniqueness of the solution to the SDE, provided that the Lipschitz continuity and linear growth conditions are satisfied. This establishes a mathematically robust foundation for the state evolution of LLMs. Additionally, we conduct a stability analysis using the Lyapunov function to ensure that, despite the inherent stochasticity, the LLM can generate stable and coherent text.

\subsection{Proof of Existence and Uniqueness}

\subsubsection{Picard-Lindelöf Theorem for SDEs}

Consider the Stochastic Differential Equation (SDE) that can be used to model the evolution of state variables in LLMs during text generation:
\[
dX(t) = \mu(X(t), t) \, dt + \sigma(X(t), t) \, dW(t)
\]
Here, \( X(t) \) represents the state of the model at time \( t \), \( \mu(X(t), t) \) is the drift term, \( \sigma(X(t), t) \) is the diffusion term, and \( W(t) \) is a Wiener process representing the stochasticity in the system.

\paragraph{Explanation:} This SDE can be interpreted as a continuous-time representation of how an LLM's internal states evolve as it generates text. The drift term \( \mu(X(t), t) \) guides the deterministic part of this evolution, while the diffusion term \( \sigma(X(t), t) \) captures the randomness inherent in the text generation process.

If the functions \( \mu \) and \( \sigma \) satisfy the Lipschitz continuity and linear growth conditions, then for any initial condition \( X(0) = X_0 \), there exists a unique solution \( X(t) \) to the SDE on the interval \( [0, T] \).

\paragraph{Explanation:} The existence and uniqueness of the solution ensure that the model's state evolution during text generation is well-defined and predictable, given the initial condition and the governing SDE.

\subsubsection{Proof Using Successive Approximations}

\paragraph{Initialization} Define the initial approximation \( X_0(t) \) as the initial condition:
\[
X_0(t) = X_0
\]
This step initializes the iterative process with the initial state of the model.

\paragraph{Successive Approximations} Define the successive approximations \( X_{n+1}(t) \) by iterating the integral form of the SDE:
\[
X_{n+1}(t) = X_0 + \int_0^t \mu(X_n(s), s) \, ds + \int_0^t \sigma(X_n(s), s) \, dW(s)
\]
Each iteration refines the approximation of the solution by incorporating the effects of drift and diffusion up to time \( t \).

\paragraph{Bounding the Differences} Using the Lipschitz condition, we can bound the difference between successive approximations:
\[
\begin{aligned}
\| X_{n+1}(t) - X_n(t) \| &\leq \int_0^t \| \mu(X_n(s), s) - \mu(X_{n-1}(s), s) \| \, ds \\
&\quad + \left| \int_0^t \| \sigma(X_n(s), s) - \sigma(X_{n-1}(s), s) \| \, dW(s) \right|
\end{aligned}
\]
The Lipschitz condition ensures that the changes in state are proportional to the differences in previous states, which is crucial for the stability of the iterative process.

\paragraph{Explanation:} By bounding these differences, we ensure that each successive approximation gets closer to the true solution, which is critical for proving the convergence of the sequence \( \{X_n(t)\} \).

\paragraph{Applying Itô Isometry} We apply Itô's isometry to handle the stochastic integral term:
\[
\begin{aligned}
E \left[ \left| \int_0^t \sigma(X_n(s), s) - \sigma(X_{n-1}(s), s) \, dW(s) \right|^2 \right] &= E \left[ \int_0^t \| \sigma(X_n(s), s) - \sigma(X_{n-1}(s), s) \|^2 \, ds \right] \\
&\leq K^2 \int_0^t E \left[ \| X_n(s) - X_{n-1}(s) \|^2 \right] \, ds
\end{aligned}
\]
Itô's isometry simplifies the treatment of the stochastic integral, converting it into a deterministic integral that is easier to analyze.

\paragraph{Combining the Bounds} Combining the deterministic and stochastic parts:
\[
E \left[ \| X_{n+1}(t) - X_n(t) \|^2 \right] \leq 2K^2 \int_0^t E \left[ \| X_n(s) - X_{n-1}(s) \|^2 \right] \, ds
\]
This bound integrates the effects of both drift and diffusion, showing how the total difference between successive approximations is controlled.

\paragraph{Explanation:} The ability to combine deterministic and stochastic bounds is key to ensuring that the iterative process is stable and converges.

\paragraph{Gronwall's Inequality} Applying Gronwall's inequality to show that \( E \left[ \| X_{n+1}(t) - X_n(t) \|^2 \right] \) converges to zero:
\[
E \left[ \| X_{n+1}(t) - X_n(t) \|^2 \right] \leq C \exp(2K^2 t) \int_0^t E \left[ \| X_n(s) - X_{n-1}(s) \|^2 \right] \, ds
\]
Gronwall's inequality is a powerful tool for proving that the sequence of approximations converges to a unique solution.

\paragraph{Convergence and Uniqueness} Thus, the sequence \( \{X_n(t)\} \) is Cauchy and converges uniformly to a limit \( X(t) \).

Finally, we show that the limit \( X(t) \) satisfies the original SDE and that no other function can satisfy the SDE with the same initial condition. This follows from the uniqueness of the limit in a complete metric space.

\paragraph{Explanation:} Convergence and uniqueness are essential to ensure that the text generation process modeled by the SDE is consistent and reliable, leading to reproducible results in the LLM's output.

\subsection{Stability Analysis for LLM Text Generation}

\subsubsection{Lyapunov Function for SDEs}

Consider the Lyapunov function \( V(X(t)) = X(t)^\top P X(t) \), where \( P \) is a positive definite matrix. For the SDE:
\[
dX(t) = \mu(X(t), t) \, dt + \sigma(X(t), t) \, dW(t)
\]
the generator \( \mathcal{L} \) applied to \( V \) is:
\[
\mathcal{L}V(X(t)) = \frac{\partial V}{\partial X} \mu(X(t), t) + \frac{1}{2} \text{Tr} \left( \sigma(X(t), t)^\top \frac{\partial^2 V}{\partial X^2} \sigma(X(t), t) \right)
\]

\paragraph{Explanation:} The Lyapunov function is used to assess the stability of the system. In the context of LLMs, this helps in understanding whether the text generation process will remain stable and produce coherent output over time.

Given \( V(X(t)) = X(t)^\top P X(t) \):
\[
\frac{\partial V}{\partial X} = 2X(t)^\top P
\]
and
\[
\frac{\partial^2 V}{\partial X^2} = 2P
\]

Thus,
\[
\mathcal{L}V(X(t)) = 2X(t)^\top P \mu(X(t), t) + \text{Tr} \left( \sigma(X(t), t)^\top P \sigma(X(t), t) \right)
\]

\paragraph{Stability Condition} To ensure stability in the text generation process, we require:
\[
2X(t)^\top P \mu(X(t), t) + \text{Tr} \left( \sigma(X(t), t)^\top P \sigma(X(t), t) \right) \leq 0
\]

If \( \mu(X(t), t) = -KX(t) \) for some positive definite matrix \( K \), then:
\[
-2X(t)^\top PK X(t) + \text{Tr} \left( \sigma(X(t), t)^\top P \sigma(X(t), t) \right) \leq 0
\]
which simplifies to:
\[
-2X(t)^\top PK X(t) + \text{Tr} \left( \sigma(X(t), t)^\top P \sigma(X(t), t) \right) \leq 0
\]

\paragraph{Explanation:} This condition ensures that the drift term (which drives the text generation) works against divergence and that the stochastic term (which introduces variability) does not overwhelm the process, leading to stable and coherent text output.

\subsubsection{Stochastic Lyapunov Functions for LLMs}

For SDEs, especially in the context of LLMs, stochastic Lyapunov functions are used, which consider the inherent randomness in the system. These functions help determine the conditions under which the generated text remains stable and coherent, either in probability or almost surely. A stochastic Lyapunov function \( V(X(t)) \) satisfies:
\[
\mathbb{E}[V(X(t))] \leq V(X(0)) + \int_0^t \mathbb{E}[\mathcal{L}V(X(s))] \, ds
\]

\paragraph{Explanation:} This inequality reflects that, on average, the system's "energy" (as captured by the Lyapunov function) does not increase uncontrollably over time, which is critical for generating coherent and contextually consistent text.

\subsubsection{Stability Condition for Stochastic Lyapunov Function}

For a stochastic Lyapunov function \( V(X(t)) \), we consider its expected value over time:
\[
\frac{d}{dt} \mathbb{E}[V(X(t))] = \mathbb{E}[\mathcal{L}V(X(t))]
\]

Given \( V(X(t)) = X(t)^\top P X(t) \):
\[
\mathbb{E}[\mathcal{L}V(X(t))] = \mathbb{E}\left[ 2X(t)^\top P \mu(X(t), t) + \text{Tr} \left( \sigma(X(t), t)^\top P \sigma(X(t), t) \right) \right]
\]

To ensure stability, we require:
\[
\mathbb{E}\left[ 2X(t)^\top P \mu(X(t), t) + \text{Tr} \left( \sigma(X(t), t)^\top P \sigma(X(t), t) \right) \right] \leq 0
\]

If \( \mu(X(t), t) = -KX(t) \), then:
\[
\mathbb{E}\left[ -2X(t)^\top PK X(t) + \text{Tr} \left( \sigma(X(t), t)^\top P \sigma(X(t), t) \right) \right] \leq 0
\]

\paragraph{Explanation:} This final condition ensures that the stochastic nature of the LLM does not destabilize the text generation process, allowing for consistent and high-quality output.

\subsection{Moment Analysis}

Moment analysis is an essential tool for understanding the behavior of Stochastic Differential Equations (SDEs) in Large Language Models (LLMs). This analysis focuses on the mean and variance of the solution \(X(t)\), and can be extended to higher-order moments, providing insights into the distributional properties of the generated text.

\subsubsection{Mean and Variance Analysis Using Itô's Lemma}

Consider the SDE governing the evolution of the state variable \(X(t)\):
\[
dX(t) = \mu(X(t), t)dt + \sigma(X(t), t)dW(t).
\]

To analyze the mean and variance of \(X(t)\), we apply Itô's lemma, which helps derive the dynamics of the first and second moments.

\paragraph{Mean Analysis}
Let \(m(t) = \mathbb{E}[X(t)]\) denote the mean of \(X(t)\). Taking the expectation of both sides of the SDE yields:
\[
\frac{d}{dt} \mathbb{E}[X(t)] = \mathbb{E}[\mu(X(t), t)].
\]
Assuming a linear relationship where \(\mu(X(t), t) = a(t)X(t)\), the mean \(m(t)\) evolves according to:
\[
\frac{d}{dt} m(t) = a(t) m(t),
\]
which solves to:
\[
m(t) = m(0) \exp\left(\int_0^t a(s) \, ds\right).
\]

\paragraph{Variance Analysis}
Let \(v(t) = \mathbb{E}[(X(t) - m(t))^2]\) represent the variance of \(X(t)\). Using Itô's lemma to find the dynamics of the second moment \(\mathbb{E}[X(t)^2]\), we obtain:
\[
\frac{d}{dt} \mathbb{E}[X(t)^2] = 2 \mathbb{E}[X(t) \mu(X(t), t)] + \mathbb{E}[\sigma^2(X(t), t)].
\]
Assuming \(\mu(X(t), t) = a(t)X(t)\) and \(\sigma(X(t), t) = b(t)\), the second moment satisfies:
\[
\frac{d}{dt} \mathbb{E}[X(t)^2] = 2a(t) \mathbb{E}[X(t)^2] + b(t)^2.
\]
From this, the variance \(v(t)\) is given by:
\[
v(t) = \mathbb{E}[X(t)^2] - m(t)^2.
\]

\subsubsection{Higher-order Moments}

The analysis can be extended to the \(n\)-th moment \(\mathbb{E}[X(t)^n]\). Using Itô's lemma, the evolution of higher-order moments is described by:
\[
\frac{d}{dt} \mathbb{E}[X(t)^n] = n \mathbb{E}[X(t)^{n-1} \mu(X(t), t)] + \frac{n(n-1)}{2} \mathbb{E}[X(t)^{n-2} \sigma^2(X(t), t)].
\]
This recursive relationship allows for the analysis of the distribution's tail behavior, which is crucial for understanding the variability and robustness of the text generated by the LLM.

\subsection{Conclusion}

The moment analysis provides a deeper understanding of the statistical properties of the SDE solutions in LLMs. By examining the mean, variance, and higher-order moments, we can infer the stability and quality of the generated text, ensuring that it remains coherent and contextually appropriate.

\paragraph{conclusion:}The mathematical proof presented in this paper shows that the text generation process of LLMs can be effectively modeled as an SDE system. The proof of existence and uniqueness guarantees that the evolution of the model’s state is well-defined and predictable under given conditions. Further stability analysis ensures that the model does not diverge during text generation, thereby producing stable and consistent output. This theoretical framework provides a crucial mathematical basis for understanding and optimizing the generative behavior of LLMs.

\section{Example: Answering the Question "What is the capital of France?"}

This section demonstrates the application of Stochastic Differential Equations (SDE) in modeling the process of answering a question using a Language Model (LLM). The specific example used is answering the question: "What is the capital of France?" with the expected response: "The capital of France is Paris."

\vspace{0.5cm}

\noindent\textbf{Step 1: Extracting Embeddings.} Using an embedding model such as BERT, each word in the question and the answer is converted into high-dimensional vectors. The question words "What," "is," "the," "capital," "of," and "France" are mapped to embeddings $\mathbf{Q}_1$, $\mathbf{Q}_2$, $\mathbf{Q}_3$, $\mathbf{Q}_4$, $\mathbf{Q}_5$, and $\mathbf{Q}_6$ respectively. The answer words "The," "capital," "of," "France," "is," and "Paris" are mapped to embeddings $\mathbf{A}_1$, $\mathbf{A}_2$, $\mathbf{A}_3$, $\mathbf{A}_4$, $\mathbf{A}_5$, and $\mathbf{A}_6$ respectively.

\vspace{0.5cm}

\noindent\textbf{Step 2: Initializing the SDE Model.} The process begins by calculating the initial state as the average of all question embeddings:
\[
\mathbf{Q}_{\text{init}} = \frac{1}{6} \sum_{i=1}^{6} \mathbf{Q}_i
\]
This initial state serves as the starting point for the SDE, which is formulated as:
\[
d\mathbf{X}(t) = \mu(\mathbf{X}(t), t)dt + \sigma(\mathbf{X}(t), t)dW(t)
\]
In this formulation, $\mu(\mathbf{X}(t), t)$ is the drift term describing deterministic trends in sequence generation, $\sigma(\mathbf{X}(t), t)$ is the diffusion term capturing stochastic fluctuations, and $dW(t)$ is the increment of the Wiener process representing random perturbations.

\vspace{0.5cm}

\noindent\textbf{Step 3: Training and Generating Answers.} The model is trained by mapping the question embeddings to the answer embeddings, enabling it to learn the dynamic transition from question to answer. The generation of the answer embeddings is carried out iteratively, starting from $\mathbf{Q}_{\text{init}}$:
\begin{align*}
\mathbf{A}_1 &= \mathbf{Q}_{\text{init}} + \mu(\mathbf{Q}_{\text{init}}, t_1) \cdot \Delta t + \sigma(\mathbf{Q}_{\text{init}}, t_1) \cdot dW(t_1) \\
\mathbf{A}_2 &= \mathbf{A}_1 + \mu(\mathbf{A}_1, t_2) \cdot \Delta t + \sigma(\mathbf{A}_1, t_2) \cdot dW(t_2) \\
\mathbf{A}_3 &= \mathbf{A}_2 + \mu(\mathbf{A}_2, t_3) \cdot \Delta t + \sigma(\mathbf{A}_2, t_3) \cdot dW(t_3) \\
\mathbf{A}_4 &= \mathbf{A}_3 + \mu(\mathbf{A}_3, t_4) \cdot \Delta t + \sigma(\mathbf{A}_3, t_4) \cdot dW(t_4) \\
\mathbf{A}_5 &= \mathbf{A}_4 + \mu(\mathbf{A}_4, t_5) \cdot \Delta t + \sigma(\mathbf{A}_4, t_5) \cdot dW(t_5) \\
\mathbf{A}_6 &= \mathbf{A}_5 + \mu(\mathbf{A}_5, t_6) \cdot \Delta t + \sigma(\mathbf{A}_5, t_6) \cdot dW(t_6)
\end{align*}

\vspace{0.5cm}

\begin{figure}[!ht]
    \centering
    \includegraphics[width=\textwidth]{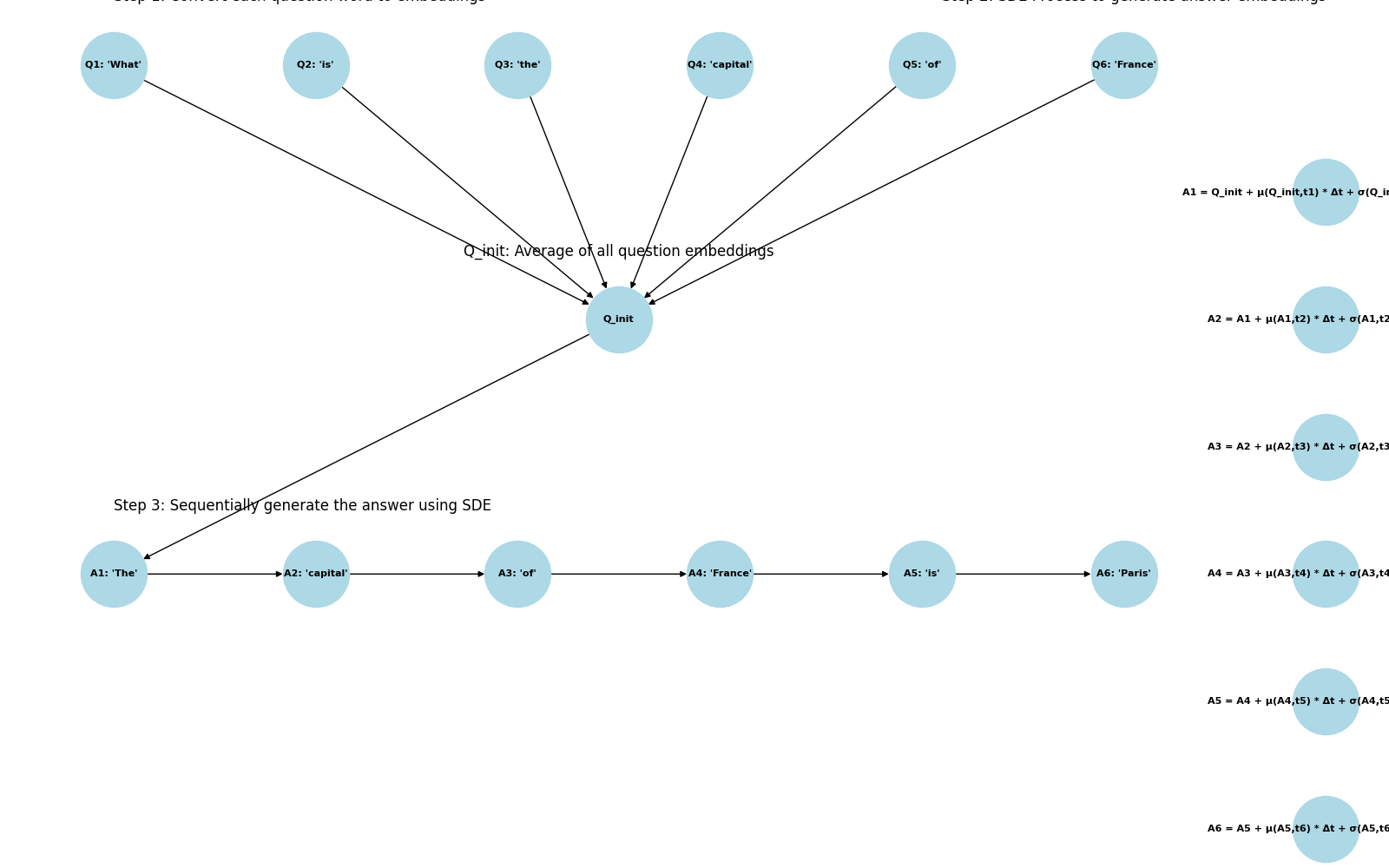}
    \caption{Illustration of the SDE-based Process for Answer Generation in LLMs. This figure demonstrates the sequential steps, starting from question embeddings, calculating the initial state, and using the SDE model to generate the final answer embeddings.}
    \label{fig:SDE_process}
\end{figure}

\end{document}